%% file: emnlp2025.tex
\documentclass[11pt]{article}

\usepackage[preprint]{acl}

\usepackage{times}
\usepackage{latexsym}

\usepackage[T1]{fontenc}
\usepackage{CJKutf8}

\usepackage[utf8]{inputenc}

\usepackage{microtype}

\usepackage{inconsolata}

\usepackage{graphicx}

\usepackage{tabularx}
\usepackage{microtype}
\usepackage{subcaption}
\usepackage{graphicx}
\usepackage{arydshln}
\usepackage{multirow}
\usepackage{multicol}
\usepackage{xcolor} 
\usepackage{colortbl}
\usepackage{booktabs}
\usepackage{tikz}
\usetikzlibrary{shapes.geometric, arrows}
\usepackage{tabularx}
\usepackage[tone]{tipa}
\usepackage{adjustbox}
\usepackage{tcolorbox}
\usepackage{geometry}
\usetikzlibrary{matrix}
\usetikzlibrary{positioning}
\usepackage{enumitem}
\usepackage[colorinlistoftodos,prependcaption,textsize=small]{todonotes}
\usepackage{hyperref}
\usepackage{makecell}
\usepackage{tipa}
\usepackage{amsmath}
\usepackage{cleveref}

\Crefname{figure}{{Fig.}}{{Figs.}}
\crefname{section}{§}{§§}
\Crefname{section}{§}{§§}
\Crefname{appendix}{{Appendix}}{{Appendices}}

\newcommand{\xhy}{\textit{xiehouyu}}

\title{Reasoning or Memorization: \\Can LLMs Understand and Generate Chinese \textit{Xiehouyu} Riddles?}

\author{
 \textbf{Hai Hu\textsuperscript{*1}},
 \textbf{Siyuan Song\textsuperscript{*2}},
 \textbf{Chongtian Shao\textsuperscript{*3}},
\\
 \textbf{Kejia Zhang\textsuperscript{3}},
 \textbf{Tianjian Zhu\textsuperscript{3}},
  \textbf{Xiaojing Zhao\textsuperscript{1}}
\\
\\
 \textsuperscript{1}The Hong Kong Polytechnic University,
 \textsuperscript{2}The University of Texas at Austin,\\
 \textsuperscript{3}Shanghai Jiao Tong University
\\
 \small{
   \textsuperscript{*}Equal Contributions
 }
 \hspace{2em}
 \small{
   \textbf{Correspondence:} \href{mailto:hai.hu@polyu.edu.hk}{hai.hu@polyu.edu.hk}
 }
}

\begin{document}
\maketitle
\input{sections/abstract}

\begin{CJK*}{UTF8}{gbsn} 
\input{sections/intro-hu}

\input{sections/categories}

\input{sections/data}

\input{sections/exp1}

\input{sections/exp2}

\input{sections/exp3}

\input{sections/related}

\input{sections/conclusion}

\section*{Limitations}
As \textit{xiehouyu} is unique in the Chinese language, and all the tested models are proficient in Chinese, our results mostly reflect their abilities in understanding this particular type Chinese riddle.  
Other languages may have other types of word games or riddles that need to be evaluated to estimate LLMs' ability more comprehensively. 

\section*{Ethics Statement}

Participants in any human validation experiments in the study were compensated with roughly 1 RMB for 1 minute, which is above the local minimum wage standard (i.e., 15 RMB for approximately 15 minutes of their time). 

\section*{Acknowledgments}

The project is funded by National Social Science Fund of China, awarded to Hai Hu (25BYY133).

\bibliography{custom}

\input{sections/appendix}

\end{CJK*}
\end{document}

%% file: sections/abstract.tex
\begin{abstract}

In this paper, we push the boundary of LLM reasoning by testing them in a Chinese language game, \xhy{}, with novel \xhy{} created by linguists that had not existed before to avoid data contamination.  
We use multiple-choice questions (MCQ), free-form explanation generation, and new \xhy{} creation to evaluate LLMs' ability to understand and create \xhy{}.
In MCQ, we use the delta of accuracy ($\Delta_{acc}$) between existing but low-frequency \xhy{} and novel ones as an index for memorization. 
$\Delta_{acc}$ for native speakers is very low, suggesting similar processing mechanisms.
However, we found that frontier Chinese models have on average a $\Delta_{acc}$ of 23.6\%, while English-centric models tested have a mean $\Delta_{acc}$ of 5.1\%, suggesting that frontier Chinese models are likely trained with much larger Chinese data, thus memorizing more low-frequency \xhy{}. 
For novel \xhy{}, Gemini 3.1 Pro demonstrated remarkable ability with acc 92.6, which is 24\% higher than human accuracy. 
In \xhy{} creation, those created by LLMs receive much worse ratings than those by humans. 
These results suggest that claims about the reasoning abilities of LLMs may need careful re-examination considering the data contamination issue, and that LLMs' creativity in language-related tasks may still be behind human experts, at least in Chinese \xhy{}. 
\end{abstract}

%% file: sections/intro-hu.tex
\section{Introduction}
\label{sec:introduction}

Non-literal meaning is an essential element of human communication.
To arrive at the correct non-literal or intended meaning,
humans, as well as Large Language Models (LLMs) need to perform various degrees of \textbf{reasoning} from the surface strings to the intended message. 
Previous work has shown that LLMs have different levels of success in such reasoning tasks, for puns~\citep{yu2018neural,xu2024good}, 
idioms~\citep{zheng2019chid,tedeschi2022id10m}, 
metaphors~\citep{neidlein2020analysis,wachowiak2023does}, 
humor~\citep{blinov2019large,hessel2023androids,jentzsch2023chatgpt} 
and conversational implicature \citep{hu2022fine,yue2024large}. 

However, whether LLMs are actually reasoning or simply repeating what they have seen in their training data is still a topic of debate~\citep{ghosh2022epic,adewumi2022potential,liu2024multilingual}, as these two mechanisms are difficult to tease apart. 
Previous work in mathematical reasoning often create uncontaminated benchmarks to tease apart memory retrieval from real reasoning~\citep{gsm-symbolic,wu2025reasoningOrMemorization-fdu}.
For language-related reasoning, studies on memorization in LLM evaluation are relatively scarce. 

\begin{table}[t]
\begin{tabular}{cll}
\multirow{4}{*}{Riddle} & 孔夫子 & 搬家 \\
 & kǒng fū zǐ & bān jiā \\
 & Confucius & move \\
 & \multicolumn{2}{l}{Confucius moving to another place} \\\addlinespace[4pt]
\multicolumn{1}{l}{} & \multicolumn{2}{l}{$\qquad\qquad\bigg\downarrow$ (semantic link)} \\\addlinespace[4pt]
\multirow{4}{*}{\begin{tabular}{l}
    Answer \\
    Literal
  \end{tabular}} & 净是 & \textcolor{blue}{书} \\
 & jìng shì & \textcolor{blue}{shū} \\
 & nothing but & books \\
 & \multicolumn{2}{l}{He has nothing but books} \\\addlinespace[4pt]
\multicolumn{1}{l}{} & \multicolumn{2}{l}{$\qquad\qquad\bigg\downarrow$ (phonetic link)} \\\addlinespace[4pt]
\multirow{4}{*}{\begin{tabular}{l}
    Answer \\
    Intended/\\
    Figurative
  \end{tabular}} & 净是 & \textcolor{blue}{输} \\
 & jìng shì & \textcolor{blue}{shū} \\
 & nothing but & lose \\
 & \multicolumn{2}{l}{I am always losing/having bad luck}
\end{tabular}
\caption{Example of a widely used \textbf{homophonic} \xhy{}, showing the reasoning steps from the `Riddle' to the \textbf{literal} and then \textbf{intended/figurative} `Answer',
involving a homophone \textit{shū} as the critical word connecting the step.
Native speakers only need to speak the `Riddle', and listeners will know the intended `Answer'. 
}
    \label{tab:xhy-example}
\end{table}

\begin{figure*}[t]
  \centering
  \includegraphics[width=0.95\linewidth]{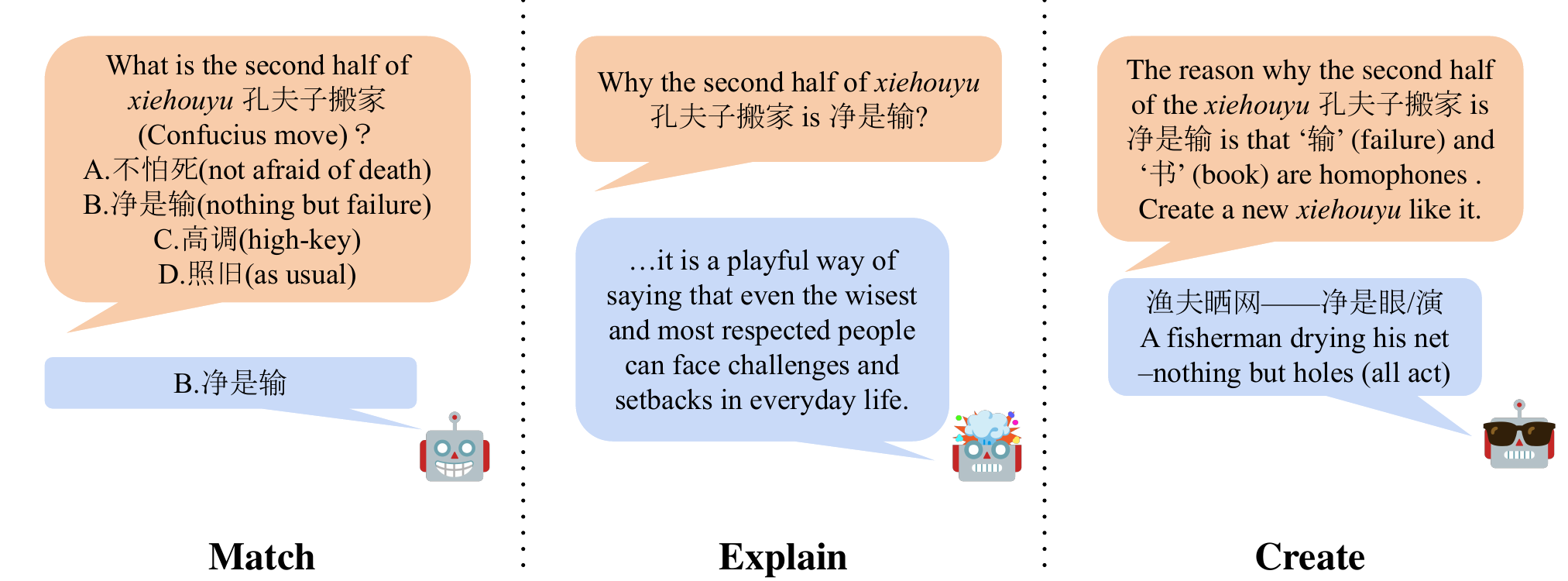}
  \caption {Overview of three experiments in this study.}
\end{figure*}

In this work, we use a special type of word game or riddle in the Chinese language---\xhy{} (歇后语)---as the testing ground for non-literal reasoning in LLMs, and design our experiments in such a way that we can potentially tell whether the model has memorized the answers or is able of complex reasoning involved in understanding \xhy{}. 
Specifically, we created hundreds of novel \xhy{} and compared model performance on these new \xhy{} and other existing ones collected from a \xhy{} dictionary.
We then use the delta of accuracy ($\Delta_{acc}$) as a measure for model memorization. 

A \xhy{} consists of two parts, a riddle, which is required, and an answer, which can and often is omitted~\citep{lai2008understanding,shu2015chinese,wang2021roles}, as shown in Table~\ref{tab:xhy-example}. 
A large proportion of \xhy{} have two \textit{answer}s, one with literal meaning (\textit{answer-lit}) and another non-literal meaning (\textit{answer-intended}).
The listener, who has heard the riddle, is expected to easily obtain the literal answer, and then reason from the literal answer to the intended answer, which often relies on word play of (1) a homophonic syllable or (2) a polysemous word that connects the literal and intended meaning. 
For instance,  in Table~\ref{tab:xhy-example}, the \xhy{} has  
`孔夫子搬家~(Confucius moving to another place)' as the riddle, 
the literal-answer: `净是书~(nothing but books)', as Confucius is an intellectual and he has nothing but books when moving, 
and finally the intended answer: `净是输~(nothing but lose)', since 书~(\textit{shū}, books) and 输~(\textit{shū}, lose) are homophones and are both pronounced as \textit{shū} in Chinese. 
This \xhy{} is commonly used when people are always losing in games or gambling.



        



\textit{Xiehouyu} as such are ideal testing ground of LLMs' reasoning abilities for intended meaning, because they involve clear reasoning steps. 
In this study, we designed three experiments of increasing difficulty to answer the following questions:

    (1) Can LLMs \textbf{match} the \textit{riddle} with the correct \textit{answer}, for existing and newly created \xhy{}? Does providing a context help? (Experiment 1)
    
    
    (2) Can LLMs \textbf{explain} the reasoning from the \textit{riddle} to the \textit{answer} in different types of \xhy{}? (Experiment 2)

    (3) 
    Can LLMs \textbf{create} reasonable and funny \xhy{}?  (Experiment 3)


To answer these questions, we created \textit{X-Riddles}, the first comprehensive dataset of 1,143 annotated \xhy{}, each comes with a riddle, a literal answer, an intended answer (if there is one), and two distractors. 
Importantly, 243 of these \xhy{} are newly created by Chinese linguists, while the other 900 existing \xhy{} are sampled from a \xhy{} dictionary, with balanced number of \xhy{} in their category (homophonic, polysemous and direct) and frequency or familiarity (low and high). 


Our results show that three frontier models perform surprisingly well on the novel \xhy{}, scoring more than 80\% in the MCQ benchmark. However, most of the Chinese frontier models have a large difference in accuracy ($\geq$ 20\%) between the low-frequency and novel \xhy{}, suggesting high level of memorization. 
LLM explanations of the \xhy{} can include wrongly identified homophonic word and sometimes exhibit over-interpretation. 
In \xhy{} creation, the two LLMs evaluated (DS-R1 and GPT-o1) fall behind humans in their reasonableness and funniness, indicating that at least in this Chinese language game, frontier models are still not as creative as humans.

%% file: sections/categories.tex
\section{Categorizations of \xhy{}}

\begin{table*}[ht]
\begin{adjustbox}{max width = \textwidth}
\begin{tabular}{llccc}
\toprule
\textbf{Type}   &\textbf{N}   & \textbf{Riddle} & \textbf{Answer-literal} & \textbf{Answer-intended/figurative} \\ \midrule
Homophony & 300 & \makecell{孔夫子搬家\\ Confucius move}  & \makecell{净是\textcolor{blue}{书}(jìng shì \textcolor{blue}{shū}) \\ nothing but \textcolor{blue}{books}   }     & \makecell{净是\textcolor{blue}{输}(jìng shì \textcolor{blue}{shū}) \\nothing but \textcolor{blue}{lose} }            \\\addlinespace[0.2cm]
Polysemy & 300 & \makecell{打开天窗 \\ open roof-window}   &   \makecell{ \textcolor{gray}{说}\textcolor{blue}{亮}\textcolor{gray}{话} \\\textcolor{gray}{to speak under }\textcolor{blue}{brightness} }  & \makecell{ 说\textcolor{blue}{亮}话 \\to speak \textcolor{blue}{frankly} }   \\\addlinespace[0.2cm] 
Other/direct   & 300 & \makecell{八仙过海 \\eight Gods cross the sea}   &   NA          & \makecell{各显神通\\ each show their super-power}            \\\addlinespace[0.2cm]
 \makecell[l]{New \\(homophony)}    & 243 & \makecell{外甥拿盾牌 \\nephew hold shield}   &   \makecell{守\textcolor{blue}{舅}(shǒu \textcolor{blue}{jiù})\\defend \textcolor{blue}{uncle}}           & \makecell{守\textcolor{blue}{旧}(shǒu \textcolor{blue}{jiù})\\defend \textcolor{blue}{old} (conservative)}            \\\addlinespace[0.2cm]

\bottomrule
\end{tabular}
\end{adjustbox}
\caption{Three types of  \textit{xiehouyu} in \textit{X-Riddles}, with their glosses. In the \textit{xiehouyu} `打开天窗——说亮话(open the roof-window —— to speak frankly)', `亮' has the literal meaning of `明亮 (bright)' and the figurative meaning of straight forward and clear. This \textit{xiehouyu} is often used when the speaker hopes for an open and honest conversation. The \textit{xiehouyu} `八仙过海——各显神通' is about the mythological story that eight gods went across the sea in different ways. This \textit{xiehouyu} is often used to describe how different people achieve success through different methods.
}
\label{tab:example-types}
\end{table*}

\paragraph{By familiarity} 
Previous work suggested that there are differences in human processing of low- and high-familiarity \xhy{}~\citep[][]{wang2021roles,qu2010mechanism,ma2019processing,zhang2013electrophysiological}.
Such differences can be explained using Standard Pragmatic Model (SPM)~\citep{grice1975logic, searle1978literal, giora1999priority}, the Direct Access Model (DAM)~\citep{gibbs1984literal, gibbs2002new} and the Graded Salience Hypothesis (GSH)~\citep{giora1997understanding}.
Specifically, the non-literal meaning of a less familiar \xhy{} must be inferred in two steps: from the \textit{riddle} to the \textit{literal} meaning and then to the \textit{non-literal} meaning.
However, it can be accessed directly from the \textit{riddle} in a more familiar \xhy{}. 
Following such intuition, we categorized collected \xhy{} into \textbf{high-familiarity} and \textbf{low-familiarity}, based on ratings from native speakers of Chinese, to see whether LLMs show different performance when processing \xhy{} of different familiarity.


\paragraph{By reasoning type}
Depending on how to derive the intended answer from the riddle, there are several types of \xhy{} as shown in Table~\ref{tab:example-types}.
The first type is \textit{homophonic}, in that one can reason from the riddle to the literal answer through semantic connection, but to arrive at the intended answer, one must identify a character or word with the same pronunciation as the one in the literal answer. 
Therefore the two characters/words share the same pronunciation, hence \textit{homophonic}.
The second type, \textit{polysemous}, hinges on a polysemous character or word, like the character 亮~(\textit{liang}) in the second row of Table~\ref{tab:xhy-example}, which can mean either ``bright'' or ``frankly''. 
Finally, some \textit{xiehouyu}, originated from two-part proverbs, legends or folklore, have a literal answer that is the same as their intended meaning. These \textit{xiehouyu} are classified as \textit{direct} in our work.

%% file: sections/data.tex
\section{Creation of \textit{X-Riddles}}
\label{sec: dataset}

In this section, we describe the creation of the \textit{X-Riddles} corpus, which contains 900 existing \xhy{} from the dictionary and 243 new \xhy{} written by us, totaling 1,143 \xhy{}, as illustrated in Table~\ref{tab:example-types}.

\paragraph{\xhy{} from dictionary}

Our \xhy{} data were originally sourced from \textit{A Comprehensive Dictionary of Chinese Xiehouyu}~\citep{wen2003chinese}, which is widely recognized in the field and contains explanations, example sentences, and other information about each \xhy{}. To construct a balanced dataset, we sampled 900 \xhy{} from three categories, 300 each category. When selecting \xhy{} expressions, we prioritized those with more example sentences in the dictionary. All selected expressions have at least one example sentence.



\paragraph{Human-created \xhy{}}
A key innovation in this study is the creation of new \xhy{} by human annotators, in order to test whether LLMs are performing actual reasoning or merely repeating their training data in understanding \xhy{}. 
We only created new homophonic \xhy{} for their scarcity (about 10\% in the dictionary) and difficulty in reasoning than the other two types of \xhy{} in our pilot study. 

There are two types of \textit{new} homophonic \xhy{} created.
The first type were created based on a \textit{seed} \xhy{}, which is an existing \xhy{} from the dictionary (N=256). 
That is, the author picks an existing homophonic \xhy{}, uses the involved homophones to create a new \xhy{}. For example, with the seed `外甥打灯笼——照旧(舅)' (A nephew holding a lantern — light the way for his uncle/same old), a linguist may use the same pair of homophones `舅(uncle)' and `旧(old)' to create a similar new \xhy{} `外甥拿盾牌——守旧(舅)' (A nephew holding a shield — guarding his uncle/sticking to old ways=conservative). 
The other type is created without any reference to existing seed \xhy{} (N=56).

Seven annotators, who are undergraduate/graduate students majoring in linguistics in a top Chinese university, were asked to create new \xhy{} from 70 seeds (at least two for each seed) and were also encouraged to write new \xhy{} without referring to seeds. 
They were also instructed to write an example sentence for each \xhy{}. 
Creating a new \xhy{} was not easy. Each annotator was given one week to create as many new \xhy{} as possible.

In the end, 312 new homophonic \xhy{} were created. 
Next, three authors of this paper independently evaluated the validity of the newly created \xhy{}. The following types of \xhy{} were identified as unreasonable: 
    (1) The \textit{riddle} does not lead to the answer, or parts of the \textit{riddle} are unnecessary for deriving the answer.
    (2) The homophonic relationship between \textit{answer-lit} and \textit{answer-intended} is incorrect.
    (3) The \textit{answer-intended} can be directly inferred from the \textit{riddle} without relying on the \textit{answer-lit}.
    (4) The \textit{answer-lit} is merely a paraphrase of the \textit{riddle}, and the homophonic component already appears in the riddle itself.

Any item judged unreasonable by two or more authors was removed. 
Finally, 243 \xhy{} are included in our dataset and used for the experiments.
Examples of the human-written \xhy{} are shown in Appendix~\ref{sec:app:examples:human:created:xhy}. 

%% file: sections/exp1.tex
\section{Experiment 1: \xhy{} matching}
\subsection{Setup}
\paragraph{Task}
In this experiment, we asked the models to choose the \textit{answer} of a \textit{riddle} from four options, one of which is the correct \textit{answer-intended}, and the other three randomly sampled from \textit{answers} of other \xhy{}. This is a standard multiple-choice question format for LLM evaluation. The prompt used is presented in Appendix~\ref{sec:app:prompt:templates}.

\paragraph{Conditions} 
Experiment 1 includes several conditions.

(1) Without-context and with-context. In the without-context condition, only the \textit{riddle} was provided; in the with-context condition, an example sentence containing the \xhy{} was presented but the \textit{answer} was replaced by empty blanks.  

(2) 0-shot, 1-shot and 3-shot prompting.
We used 0-shot prompting multiple-choice questions and asked model to give answers in letter form (`A', `B', `C' or `D'). 

(3) Chain-of-Thought (CoT) and no CoT~\citep{wei2022chain}.

(4) Literal answer and figurative answer. For instance, whether the answer is ``nothing but books'' or ``nothing but lose'' in the Confucius move example. We conjecture that figurative answer will be much harder for the models because it cannot use shallow surface cues of the literal answer. Note that in the human experiment, we are using the figurative answer. 

Model performance is measured by their accuracy in answering the questions, with random guessing at 25\% accuracy.


\paragraph{Models}
We tested a wide range of language models, including the Qwen2.5-Instruct~\citep{qwen2.5} model family (ranging from 0.5B-72B), 
and frontier reasoning models, some with Chinese origin, mostly open-weights: DeepSeek-V3.2~\citep{deepseekai2025deepseekr1}, Kimi-k2.5~\citep{kimi-k2}, Minimax-m2.5, GLM-5, Doubao-Seed-2.0, and other English-centric models from the US, mostly close-sourced: Gemini-3.1-Pro, GPT-5.2, Grok-4, Claude-Opus-4.6, among others.\footnote{Qwen-2.5 models are evaluated locally on University cluster. DeepSeek-V3.2, Qwen3.5-plus and Kimi-k2.5 are accessed via their official API in Feb 2026. All other models are accessed via Openrouter in March 2026.}

We first evaluated Qwen2.5-0.5B to 72B, Qwen3.5-Plus and DeepSeek-V3.2 on all conditions mentioned above, to investigate the impact of different conditions (Sections~\ref{sec:exp1:overall:acc} and \ref{sec:exp1:conditions}). 
Based on the results, all other frontier models were only evaluated on a single condition (CoT=yes, nshot=0, context=no, LorF=Fig) for homophonic \xhy{} (Section~\ref{sec:exp1:reasoning:or:memorization}). 
We selected this condition because it is the best one to evaluate the true reasoning ability of language models. 
Only homophonic \xhy{} were evaluated for these models because we observe that the other two types of \xhy{} (polysemous and direct) were too easy for frontier models. 


\subsection{Human accuracy on X-Riddles}

To estimate how good humans are in choosing the correct answers to the riddles, we asked more than 100 native speakers of Chinese to do the MCQ task first. 
That is, they were asked to choose the most likely \textit{answer} given a \textit{riddle}, out of four choices. We also instructed them to rate their familiarity of the \xhy{} on a scale of 1 to 4, with 1 being ``have not seen the \xhy{} at all'' and 4 being ``very familiar with the \xhy{}''.

All \xhy{} in \textit{X-Riddles} were randomly split into 17 lists, each containing approximately 70 riddles in the form of multiple-choice questions, plus two catch trials. Participants were randomly assigned to one list, and those who answered the catch trials wrong were excluded from the final analysis. Participants were compensated monetarily  for completing the task, which took about 20 minutes. In the end, each \xhy{} was rated by at least five native speakers. 


We make several observations from the results presented in in Table~\ref{tab:human:results}.

\paragraph{High-familiarity \xhy{} has high accuracy, likely accessed via memory retrieval. } As expected, native speakers achieved near-perfect accuracy for \textit{xiehouyu} they were highly familiar with, scoring 99.0\%, 96.9\%, and 95.9\% for the homophony, other, and polysemous categories, respectively. This indicates that when the \xhy{} is known, the task is trivial and relies on memory recall. 

\paragraph{Low-familiarity and novel \xhy{} have similar accuracy, suggesting similar reasoning mechanisms. } 
The lowest performance among existing \xhy{} was on low-familiarity homophonic \textit{xiehouyu} (71.5\%). \textbf{Crucially, this is very close to the accuracy on the newly created homophonic items (68.6\%). }
We take this finding to mean that for humans, processing low-frequency and novel homophonoic \xhy{} involves the \textit{same} mechanism; that is, both require \textit{reasoning} rather than memory retrieval. This is in sharp contrast to how LLMs (especially DeepSeek-R1) processes the two categories of \xhy{}, as we will see later. 

\paragraph{Homophonic \xhy{} is the hardest. } 
A more detailed analysis of the low-familiarity and novel items reveals \textbf{varying levels of difficulty across different types of \xhy{}}. Notably, participants performed best on unfamiliar ``polysemous'' (83.3\% accuracy) and ``other'' types (81.8\%), suggesting that the semantic or logical links in these categories are more transparent or easier to reason about than phonetic ones. 

These results establish a robust benchmark for human-level reasoning on this task, highlighting that 
resolving polysemous puns is more intuitive for humans than resolving those based on homophony.

\begin{table}[t!]
    \centering
    \small
\begin{tabular}{lcccc}
\toprule
Type & Familiarity & Acc. & $n$ \\
\midrule
Homophony & high & 99.0 & 12 \\
Homophony & low & 71.5 & 288 \\\midrule
New & - & 68.6 & 242 \\\midrule
Other & high & 96.9 & 61 \\
Other & low & 81.8 & 239 \\\midrule
Polysemy & high & 95.9 & 25 \\
Polysemy & low & 83.3 & 275 \\\midrule
Total & & & 1,142\\
\bottomrule
\end{tabular}
    \caption{Accuracy of native speakers on MCQs. 
    \xhy{} with a mean familiarity $\geq2.5$ (Likert-scale from 1 to 4) is categorized as ``high'' familiarity.
    \textit{n} refers to the number of \xhy{} in the category.  }
    \label{tab:human:results}
\end{table}

\subsection{LLMs' accuracy on X-Riddles}

\begin{figure*}[t!]
    \centering
    \includegraphics[width=0.99\textwidth]{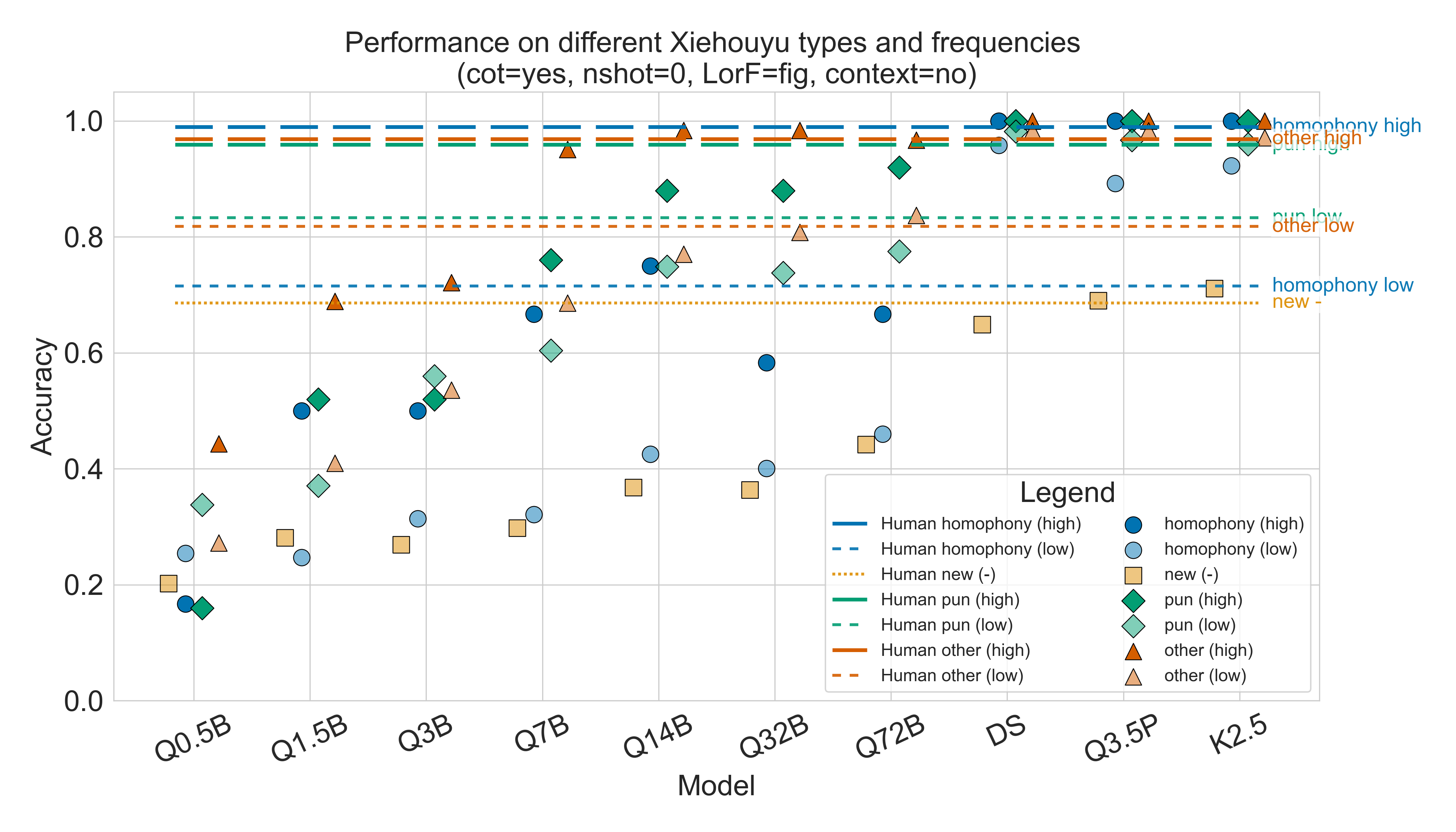}
    \caption{Results for experiment 1: accuracy of \textbf{0-shot} multiple-choice question for \xhy{} for LLMs and humans. Different shapes or colors refer to different types of \xhy{}. The lines represent human accuracy. }
    \label{fig:exp1:0shot}
\end{figure*}







\subsubsection{Overall accuracy} \label{sec:exp1:overall:acc}
The overall accuracy for all models in the 0-shot setting is presented in Figure~\ref{fig:exp1:0shot}. 
We first observe scaling effects for the Qwen2.5 series, demonstrated by the consistent increase of model accuracy as model size increases from 0.5B to 72B. 
While Qwen2.5-0.5B performs at near chance level (25\%), the larger Qwen models demonstrate high proficiency, especially in \textit{polysemous} and \textit{other} types of \xhy{} of high-familiarity to humans. 
Frontier models (DeepSeek-V3.2, Qwen3.5-Plus, Kimi-2.5) have almost perfect accuracy for polysemous and other \xhy{}, as well as homophonic \xhy{} with high familiarity/frequency.


\subsubsection{Effects of evaluation conditions.}
\label{sec:exp1:conditions}

Figure~\ref{fig:exp1:eval:conditions} shows the effects of other conditions in the accuracy of experiment 1. 
Chain-of-Thought prompting (blue bars) hurts the performance for Qwen2.5-1.5B and Qwen2.5-3B, and has no noticable influence on other models. 
For few-shot prompting (orange bars), increasing from 0-shot to 3-shot increases the performance for Qwen2.5 models for at most 4 percentage points, but almost no effect for large MoE models such as Qwen3.5-Plus and DeepSeek-V3.2. 
Providing context (green bars) or using the literal answer (red bars) are much more helpful, with up to 10\% improvement in performance, but these are not the settings in our human experiment, and thus not reported in later sections of the paper. 
Based on these piloting results, we use CoT=yes, n-shot=0, no context and figurative answer as the evaluation condition for all frontier models next.

\subsubsection{Reasoning or Memorization?}
\label{sec:exp1:reasoning:or:memorization}

Our most important analysis lies in whether the models are performing true reasoning or have simply memorized the \xhy{} from their training data.
We use the difference of performance on low-familiarity homophonic \xhy{} and the new ones ($\Delta$ between Low and New) as the measure, shown in the last column of Table~\ref{tab:homo:full:results}. 

\textbf{Chinese models seem to memorize much more than English-centric models, evidenced by large $\Delta_{acc}$.}
For humans, $\Delta$ is very low (2.9), suggesting that when processing \xhy{} of low-familiarity and new \xhy{}, humans rely on the same mechanism, that is genuine reasoning. 
Frontier Chinese models stood out by having (extremely) large $\Delta$, 23.6 on average. 
DeepSeek-V3.2, for instance, has a $\Delta$ of 30.9, with almost perfect accuracy on low-familiarity ones (95.8\%).
This result suggests that Chinese models may have been trained on much more Chinese data, which contains existing but very rare \xhy{}. 
However, English-centric models have rather low or even negative $\Delta$, only 5.1 on average (below 10 for all, and even -2.8 for Gemini 3.1 Pro), suggesting that they were probably not trained on massive Chinese data that may contain rare \xhy{}.
This is understandable since the Chinese labs have likely put more effort in obtaining large-scale Chinese data to train their models.

\input{tables/exp1_homo_acc}

\paragraph{Token usage more than doubled for new \xhy{} in Chinese models.}
Table~\ref{tab:token-compact} presents the increase in token usage (response + thinking) for each model under from High to Low to New, for homophonic \xhy{}.
We again observe a difference among frontier Chinese vs.~English-centric models. 
There is a much larger token increase from High to Low in Chinese models compared to English-centric ones (91.8\% vs.~36.2\%) and from High to Low (166.3\% vs.~38.8\%). 
Notice again that for Chinese models there is a 40\% increase from Low to New, but the increase is only 1.8\% for English-centric models, suggesting that the Low and New \xhy{} are treated differently by Chinese models, but not English-centric models. 
These results reflect that the Chinese models are trying very hard to reason about the Low and New \xhy{}. However, their performance on the New did not increase as their thinking effort increases.

\input{tables/table_token_compact}

\paragraph{Overthinking does not lead to improved performance. } 
Chinese frontier models show a striking dissociation between effort and performance on New items: they produce on average 40.6\% more tokens compared to Low items, yet their accuracy drops by 23.6 percentage points. This pattern is consistent with the ``overthinking'' phenomenon documented in
\citet{chen2024-overthink-o1,sui2025stop-overthinking}, where extended reasoning chains fail to improve and may even degrade performance.
In contrast, English-centric models show minimal token increase (1.8\%) alongside minimal accuracy change, suggesting a more uniform processing strategy across Low and New items. Notably, the best-performing model on New items (Gemini 3.1 Pro) actually shows a slight decrease in tokens from Low to New (-4.4\%), consistent with findings that more capable models allocate computational resources more efficiently~\citep{lin2025plan-and-budget}.


\subsubsection{Discussion}

The key interpretive tool in Experiment 1 is $\Delta_{acc}$, the accuracy gap between Low-familiarity and New items.
Because the human baseline shows a near-zero $\Delta_{acc}$, any large gap observed in a model is difficult to attribute to differences in item difficulty and instead points to memorization of existing \xhy{} from training data.
This diagnostic reveals a clear split: Chinese-origin models exhibit substantial $\Delta_{acc}$ values, while English-centric models do not---though we note that the two groups also differ in openness and scale, making it difficult to fully isolate the role of Chinese training data volume.

Gemini 3.1 Pro is particularly noteworthy: it surpasses human accuracy on New items while maintaining a near-zero $\Delta_{acc}$, suggesting that genuine multi-step reasoning---from riddle to literal answer to phonetic link---is within reach of frontier models.
More generally, $\Delta_{acc}$ provides only a \textit{lower bound} on memorization: a model that has both memorized existing items and learned to reason will still appear reasoning-driven by this metric.

These observations carry implications beyond \xhy{}.
Any benchmark built from culturally circulated material risks conflating retrieval with reasoning, and the effect is amplified for languages well represented in training corpora.
Our design---pairing existing items with newly created ones of matched difficulty---offers a general template for disentangling the two, which we believe deserves wider adoption.
In Experiment~2, we move beyond answer selection to examine whether models can \textit{explain} the reasoning steps underlying \xhy{}.

%% file: tables/exp1_homo_acc.tex
\begin{table}[t]
\centering\small
\resizebox{\columnwidth}{!}{%
\begin{tabular}{@{}l rrrr@{}}
\toprule
\textbf{Model} & \textbf{High} & \textbf{Low} & \textbf{New} & $\boldsymbol{\Delta}$ \\
\midrule
\multicolumn{5}{@{}l}{\textit{Human Baseline}} \\
Human & 99.0 & 71.5 & 68.6 & 2.9 \\
\midrule
\multicolumn{5}{@{}l}{\textit{Qwen2.5 (Open-Weights)}} \\
Qwen2.5-0.5B & 16.7 & 25.7 & 20.2 & 5.5 \\
Qwen2.5-1.5B & 50.0 & 24.6 & 28.1 & \(-\)3.5 \\
Qwen2.5-3B & 50.0 & 31.6 & 26.9 & 4.7 \\
Qwen2.5-7B & 66.7 & 32.3 & 29.8 & 2.5 \\
Qwen2.5-14B & 75.0 & 42.7 & 36.8 & 5.9 \\
Qwen2.5-32B & 58.3 & 40.3 & 36.4 & 3.9 \\
Qwen2.5-72B & 66.7 & 46.2 & 44.2 & 2.0 \\
\rowcolor[gray]{0.93}
\textit{Mean (Qwen2.5)} & \textit{54.8} & \textit{34.8} & \textit{31.8} & \textit{3.0} \\
\midrule
\multicolumn{5}{@{}l}{\textit{Frontier Chinese Models}} \\
DeepSeek-V3.2 & 100.0 & \textbf{95.8} & 64.9 & 30.9 \\
Doubao-Seed2.0 & 100.0 & 95.5 & 81.8 & 13.7 \\
GLM-5 & 100.0 & 86.8 & 70.2 & 16.6 \\
Kimi-k2.5 & 100.0 & 92.4 & 71.1 & 21.3 \\
MiniMax-m2.5 & 100.0 & 83.7 & 45.0 & 38.6 \\
Qwen3.5-Plus & 100.0 & 89.2 & 69.0 & 20.2 \\
\rowcolor[gray]{0.93}
\textit{Mean (Chinese)} & \textit{100.0} & \textit{90.6} & \textit{67.0} & \textit{23.6} \\
\midrule
\multicolumn{5}{@{}l}{\textit{Frontier English-centric Models}} \\
Claude 4.6 Opus & 100.0 & 85.1 & 83.9 & 1.2 \\
Claude 4.6 Sonnet & 91.7 & 70.5 & 67.8 & 2.7 \\
Gemini 3.1 Pro & 100.0 & 92.4 & \textbf{92.6} & \(-\)0.2 \\
Gemini 3 Flash & 100.0 & 87.8 & 77.7 & 10.2 \\
GPT-5.2 & 100.0 & 80.2 & 72.3 & 7.9 \\
Grok-4$^\dagger$ & 100.0 & 80.2 & 71.1 & 9.1 \\
\rowcolor[gray]{0.93}
\textit{Mean (Eng.-centric)} & \textit{98.6} & \textit{82.7} & \textit{77.5} & \textit{5.1} \\
\bottomrule
\end{tabular}
}
\caption{Accuracy (\%) on homophonic \textit{xiehouyu} across High Familiarity, Low Familiarity, and New (uncontaminated) splits. \(\Delta_{acc}\) represents the performance drop from Low to New. \textbf{Frontier Chinese models overall show much more memorization (\(\Delta_{acc}\) as high as 38.6), compared to English-centric Models (\(\Delta_{acc}\) less than 10), which may be attributed to the potentially much more Chinese training data.} Furthermore, while the best performing models for New \xhy{} are Close-Weights, some Open-Weights models (e.g., Kimi-2.5) reach very high accuracy.}
\label{tab:homo:full:results}
\end{table}

%% file: tables/table_token_compact.tex
\begin{table}[t]
\centering\small
\begin{tabular}{@{}l r r r r@{}}
\toprule
& & \multicolumn{3}{c}{\textbf{Total Token $\Delta$\%}} \\
\cmidrule(l){3-5}
\textbf{Model} & $\boldsymbol{\Delta_{\textbf{Acc}}}$ & \textbf{L\,/\,H} & \textbf{N\,/\,H} & \textbf{N\,/\,L} \\
\midrule
\multicolumn{5}{@{}l}{\textit{Frontier Chinese Models}} \\
DeepSeek-V3.2 & 30.9 & +71.6 & +220.6 & +86.8 \\
Doubao-Seed2.0 & 13.7 & +61.1 & +117.2 & +34.8 \\
GLM-5 & 16.6 & +98.4 & +148.6 & +25.3 \\
Kimi-k2.5 & 21.3 & +98.6 & +197.3 & +49.7 \\
MiniMax-m2.5 & 38.6 & +149.4 & +190.2 & +16.3 \\
Qwen3.5-Plus & 20.2 & +71.5 & +123.7 & +30.4 \\
\rowcolor[gray]{0.93}
\textit{Mean (Chinese)} & \textit{23.6} & \textit{+91.8} & \textit{+166.3} & \textit{+40.6} \\
\midrule
\multicolumn{5}{@{}l}{\textit{Frontier English-centric Models}} \\
Claude 4.6 Opus & 1.2 & +67.3 & +76.0 & +5.2 \\
Claude 4.6 Sonnet & 2.7 & +49.1 & +51.8 & +1.8 \\
Gemini 3.1 Pro & \(-\)0.2 & +18.4 & +13.2 & \(-\)4.4 \\
Gemini 3 Flash & 10.2 & +21.8 & +30.5 & +7.2 \\
GPT-5.2 & 7.9 & +32.7 & +33.9 & +0.9 \\
Grok-4$^\dagger$ & 9.1 & +27.8 & +27.7 & \(-\)0.1 \\
\rowcolor[gray]{0.93}
\textit{Mean (Eng.-centric)} & \textit{5.2} & \textit{+36.2} & \textit{+38.8} & \textit{+1.8} \\
\bottomrule
\end{tabular}
\\[4pt]
\caption{Token effort analysis across item types for frontier models.
$\Delta_{\text{Acc}}$ = Acc(Low)\,$-$\,Acc(New) in percentage points.
Token $\Delta$\% columns show percentage change in total tokens
(response\,+\,thinking) relative to the reference split. For instance, L/H shows increase in token count from Low to High. {\footnotesize $^\dagger$\,No extended thinking; total tokens $=$ response tokens only.}}
\label{tab:token-compact}
\end{table}

%% file: sections/exp2.tex
\section{Experiment 2: \xhy{} explanation}
\subsection{Experiment Setup}
\paragraph{Task}
In this experiment, the models are asked to explain why the intended meaning of a \textit{riddle} is the \textit{answer-intended}. We directly prompt the model to provide an explanation. 

Our aim is to see whether the models truly ``understand'' the underlying logic of \xhy{} when matching \textit{riddle} and \textit{answer}. 
To do so, we sampled a total of 432 (144 $\times$ 3) model explanations of \xhy{} and analyzed them closely to see if the responses show real understanding.
The three models are: DeepSeek-R1, Qwen-max and  Kimi-k2.\footnote{These models were tested in July 2025.}
The prompts can be found in Appendix~\ref{sec:app:prompt:templates}.

\begin{table}[h!]
    \centering
    \begin{tabular}{cccc}\toprule
       Type & N & Type & N \\\midrule
       homophony-high  & 12 & homophony-low & 12 \\
       pun-high  & 20 & pun-low & 20 \\
       other-high  & 20 & other-low & 20 \\
       new  & 40 &&  \\\bottomrule
    \end{tabular}
    \caption{A total of 144 \xhy{} sampled to obtain model explanation: N refers to numbers in each category.}
    \label{tab:exp2:sampled:numbers}
\end{table}

\paragraph{Evaluation}
We performed two rounds of analyses. In the first round, we asked 56 native speakers of Chinese to rate LLMs' explanations on three aspects: (1) whether the explanation mentions the homophonic word (Yes/No/NA because there is no homophonic word), (2) whether (they think) the LLMs have truly understood the \xhy{} (Yes/No/Uncertain) and (3) the quality of the explanations on a Likert-scale of 1 to 5. 
In the second round, the authors selected all explanations with an average quality score no higher than 2 (about 30 explanations) and sampled another 30 explanations with a quality score of 5, and manually went through them to identify the sensible and nonsensical segments.

\subsection{Results}

\begin{table}[t]
    \centering
    \small
    \begin{tabular}{ccccc}
        \toprule
        \textbf{Model} & \textbf{Quality} & \textbf{Acc} & \textbf{Understd.} & \textbf{Res. Len.} \\
        \midrule
        DS-R1 & 4.11$\pm$0.93 & 0.78 & 0.82 & 802.41 \\ 
        Q-Max & 3.92$\pm$1.21 & 0.64 & 0.73 & 897.91 \\
        Kimi-k2 & 3.83$\pm$1.17 & 0.70 & 0.71 & 517.71 \\
        \bottomrule
    \end{tabular}
        \caption{Ratings of model explanations on \xhy{}. \textit{Acc} refers to the accuracy of identifying the homophonic word. \textit{res len}  refers to the length of response. }
    \label{tab:exp2:results}
\end{table}

\paragraph{Performance of Models} The ratings of model explanations on \xhy{} by native speakers are as shown in Table~\ref{tab:exp2:results}. We observe that DeepSeek-R1 achieves the highest quality and understanding-ratio, while Qwen-max ranks second, producing slightly longer responses. Kimi-k2 demonstrates inferior performance, with responses markedly shorter than those of the other two models. An Ordinary Least Squares (OLS) regression, \texttt{quality\_score $\sim$ res\_len}, shows no statistically significant association between response length and quality score, suggesting that the observed differences are not merely driven by verbosity. In addition, although the models can choose the correct answer for the majority of homophonic \xhy{}, as shown in the results of Experiment 1 in Table~\ref{tab:homo:full:results}, their accuracy in identifying the homophonic word tend to be worse as shown in Table~\ref{tab:homo-acc}, especially for novel \xhy{}, indicating a lack of phonological knowledge. 

\begin{table}[h!]
\centering
\small
\begin{tabular}{lccccc}
\toprule
\textbf{Model \textbackslash  ~Familiar.} & \textbf{New} & \textbf{Low} & \textbf{High}\\
\midrule
DS-R1 & 85.0 & 83.4 & 100.0 \\
Q-max & 67.5 & 45.4 & 100.0  \\
Kimi-k2 & 67.5 & 100.0 & 91.7\\
\bottomrule
\end{tabular}
\caption{Ratings on whether the models accurately in identifying the homophonic word (if any) with respect to familiarity.}
\label{tab:homo-acc}
\end{table}

\begin{figure*}[t]
  \centering
  \includegraphics[width=0.95\linewidth]{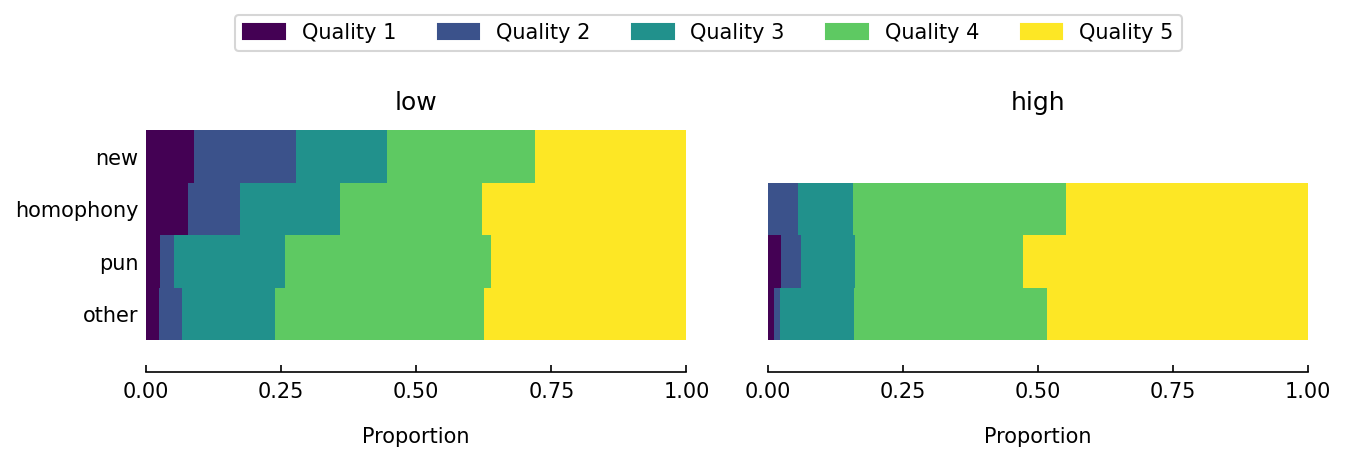}
  \caption {Quality distributions across models and \xhy{} types and familiarity.}
  \label{fig:expr2-quality}
\end{figure*}

\paragraph{The Effect of Different \xhy{} Types} The distributions of the quality scores of the three models across \xhy{} types and familiarity are illustrated in Figure~\ref{fig:expr2-quality}. The majority of explanations for high-familiarity \xhy{} receive a quality above 4.0 (yellow and light green), while those for low-familiarity ones are considerably worse (although still acceptable), which holds for all three models. 
As for novel \xhy{}, the explanation quality shows a clear degradation, especially for Kimi-k2 and Qwen-max, in which only $\sim$50\% of their explanations are rated 4.0 or higher. 

Taken together, the results suggest that for highly familiar \xhy{}, the best models on the market in general provide sensible explanations. However, for newly written \xhy{}, only half of their responses are rated as high quality. 


\paragraph{Models' weaknesses in understanding \xhy{}.}\label{sec:exp2:weakness}
In the second round of analysis, we performed manual inspection of model explanations and identified the following types of major errors. 

First, models failed to identify the homophonic character or word, or they find the wrong one. 
A careful examination shows that the former is the majority. 
This usually occurs when the homophone is not a strict one, such as ``寺'' (\textit{si}) and ``事'' (\textit{shi}), which sound similar in some varieties of Mandarin, and are treated as homophone in many \xhy{}, but are nonetheless not exactly the same in Standard Mandarin.\footnote{The full IPA for two characters in Standard Mandarin: /\textipa{s\s{\*r}}/ and /\textipa{\:s\s{\textturnrrtail}}/. However, in many southern varieties of Mandarin, /s/ (alveolar fricative) and /\textipa{\:s}/ (retroflex fricative) are not distinguished. The vowels are similar too. Note that the Pinyin for the vowels are the same, but they are different vowels.
}
The latter arises when the model seems to be trying too hard to find the homophonic word. When no clear homophonic word is identified, the model will conjure up the homophone and usually claim, incorrectly, that in some Mandarin dialect or colloquial speech such a homophone exists, as shown in Table~\ref{tab:homophone_errors} in the Appendix.
Such behavior has been reported previously where models hallucinate or produce wrong reasoning~\citep{turpin2023language-models-dont-say-what-they-think}.



Two, models are prone to over-interpretation and even hallucination. 
Table~\ref{tab:exp2:over-interpretation:example} in the Appendix shows an example of over-interpretation, which happened most often because models failed to identify the homophonic words.

%% file: sections/exp3.tex
\section{Experiment 3: \xhy{} creating}

\begin{table*}[ht]
\centering
\begin{adjustbox}{max width = \textwidth}
\begin{tabular}{@{}lllllll@{}}
\toprule
\textbf{Author} & \textbf{Riddle} & \textbf{Answer-lit} & \textbf{Answer-intended} & \textbf{Reasonable} & \textbf{Funny} & \textbf{Overlap} \\ \midrule

GPT-o1 & 
\makecell[l]{在山洞里放爆竹 \\ Setting off firecrackers in a cave} & 
\makecell[l]{\textcolor{blue}{响}得深 \\ \textcolor{blue}{Resound} deeply} & 
\makecell[l]{\textcolor{blue}{想}得深 \\ \textcolor{blue}{Think} deeply} & 
\makecell[l]{1} & 
\makecell[l]{3.00} & 
\makecell[l]{0.27}\\ \addlinespace

GPT-o1 & 
\makecell[l]{水壶烧开冒白烟 \\ Boiling kettle emitting white smoke} & 
\makecell[l]{\textcolor{blue}{蒸汽} \\ \textcolor{blue}{Steam}} & 
\makecell[l]{\textcolor{blue}{争气} \\ \textcolor{blue}{Strive} for success} & 
\makecell[l]{1} & 
\makecell[l]{2.67} & 
\makecell[l]{0.29}\\ \addlinespace

DS-R1 & 
\makecell[l]{膝盖上钉掌 \\ Horseshoe on knee} & 
\makecell[l]{离\textcolor{blue}{蹄} \\ Away from \textcolor{blue}{hoof}} & 
\makecell[l]{离\textcolor{blue}{题} \\ Off \textcolor{blue}{topic}} & 
\makecell[l]{1} & 
\makecell[l]{2.67} & 
\makecell[l]{0.33}\\ \addlinespace


DS-R1 & 
\makecell[l]{黄豆配绿豆 \\ Soy bean with mung bean} & 
\makecell[l]{双\textcolor{blue}{豆}\\ Double \textcolor{blue}{bean}} & 
\makecell[l]{双\textcolor{blue}{斗} \\ Double \textcolor{blue}{fight}} &
\makecell[l]{0} & 
\makecell[l]{1.00} & 
\makecell[l]{0.27}\\ \addlinespace

Human & 
\makecell[l]{和尚流浪 \\ Wandering monk} & 
\makecell[l]{没\textcolor{blue}{寺}找\textcolor{blue}{寺} \\ No \textcolor{blue}{temple}, seek \textcolor{blue}{temple}} & 
\makecell[l]{没\textcolor{blue}{事}找\textcolor{blue}{事} \\ Ask for \textcolor{blue}{trouble}} & 
\makecell[l]{1} & 
\makecell[l]{4.00} & 
\makecell[l]{0.38}\\ \addlinespace

Human & 
\makecell[l]{猪八戒的学生 \\ Zhubajie's student} & 
\makecell[l]{\textcolor{blue}{悟能}之徒 \\ Disciple of \textcolor{blue}{Wuneng}} & 
\makecell[l]{\textcolor{blue}{无能}之徒 \\ \textcolor{blue}{Incompetent} person}  & 
\makecell[l]{1} & 
\makecell[l]{3.67} & 
\makecell[l]{0.47}\\ \addlinespace

\bottomrule
\end{tabular}
\end{adjustbox}
\caption{Examples of model- and human-created \xhy{}. Blue text highlights homophonic puns, where words share the same (or similar) sounds but carry different meanings. In general, humans tend to follow traditional patterns that rely on folklore and vulgar humor, whereas LLMs leverage physical knowledge to construct novel \xhy{}.}
\label{tab:exp3_example_comparison}
\end{table*}

\subsection{Experimental setup}
\paragraph{Task}
In this experiment, the models were asked to create new \xhy{}. 
Since previous sections show that homophonic \xhy{} are most difficult and we already have human-created homophonic \xhy{}, we focus on homophonic ones in this experiment. 
In our pilot experiment, we found it very difficult to elicit models to create high quality \xhy{} without giving example \xhy{} and the logic behind them. 
Therefore, in the prompt
we provide an existing \xhy{}$_1$, its detailed reasoning steps, \textit{and} a human-created \xhy{}$_2$, which uses \xhy{}$_1$ as the seed.
The models were not forced to follow the seed strictly by using the same homophones, but were prompted to create a ``similar'' \xhy{}. 
The complete prompt is shown in Appendix~\ref{sec:app:prompt:templates}.

Twenty seed \xhy{} were used and each model was prompted three times for one \xhy{}.
Therefore, each model generated 60 new \xhy{}.

\paragraph{Evaluation}
Three authors evaluated the \xhy{} created by models in two dimensions: reasonableness and funniness. For reasonableness, they were asked to judge whether a new \xhy{} is reasonable (1) or not (0), following same metrics for human \xhy{} validation (\Cref{sec: dataset}). If deemed reasonable, the \xhy{} was then rated on a scale of 1 to 3. For funniness, they were asked to give a score ranges from 1 (not funny at all) to 4 (incredibly funny).
We also compute a novelty score based on Jaccard index to see whether the \xhy{} has significant overlap with existing ones from the dictionary, to make sure that the models have created genuine novel ones (see Appendix~\ref{sec:app:novelty:measure}).

\paragraph{Models}
We performed a pilot study with several models, including GPT4o, Qwen2.5-72B, o1, and DeepSeek-R1. 
We observed that only o1 and DeepSeek-R1 produced valid \xhy{}. Therefore, we conducted evaluation on these two models.

\subsection{Results}
\begin{table}[h!]
    \centering
    \small
        \begin{tabular}{lcccc}
        \toprule
        \textbf{Src.} & \textbf{N} & \textbf{Reasonable} & \textbf{Funniness} & \textbf{Max Ovrlp} \\
        \midrule
        Human & 312 & 0.75 (0.34) & 2.26 (0.74) & 0.32 \\
        DS-R1 & 60 & 0.43 (0.40) & 1.47 (0.54) & 0.30 \\
        o1 & 60 & 0.47 (0.37) & 1.62 (0.60) & 0.29 \\
        \bottomrule
    \end{tabular}

        \caption{Performance comparison of human and model-generated \xhy{}. N refers to numbers of \xhy{}. Scores are reported as Mean (Standard Deviation).}
    \label{tab:exp3_performance_comparison}
\end{table}

\paragraph{Human and Model Performance}
Table~\ref{tab:exp3_performance_comparison} presents the ratings for \xhy{} generated by humans, DeepSeek-R1 and o1. Human-generated \xhy{} were rated higher than both models in reasonableness and funniness, though LLMs exhibited marginally lower overlap with existing examples. Comparing the two models, o1 performed slightly better than DeepSeek-R1. These results suggest that while LLMs can produce somewhat reasonable and funny \xhy{}, they have not yet surpassed original human creativity. 

\paragraph{Strengths and Weaknesses of Models in Generating \xhy{}}
While LLMs show promise, they currently lag behind humans in crafting reasonable and humorous \xhy{}. The primary bottleneck lies in context integration. Models frequently bypass logical reasoning to fabricate connections between the riddle and the answer. This tendency to over-explain or hallucinate relationships aligns with our findings in Section~\ref{sec:exp2:weakness}, where models failed to grasp homophony and low-frequency examples due to the lack of phonetic information. 

In terms of humor, human-written \xhy{} were more dynamic and versatile, while model-generated ones tended to be mediocre and safe. 
On the other hand, both DS-R1 and o1 showed slightly lower overlap scores than humans, although the difference was very small.

\paragraph{Example \xhy{} generated by LLMs.}
Table~\ref{tab:exp3_example_comparison} demonstrates the riddles generated by LLMs and humans. 

Although humans remain more versatile in humor, LLMs successfully leverage physical knowledge to construct novel \xhy{}. Unlike traditional \xhy{}, which typically rely on folklore or vulgar humor, LLMs demonstrate a capacity for broader creativity by incorporating concepts from modern physics and general knowledge. However, relying on such common knowledge may result in less humor, as the core of \xhy{} often depends on specific cultural contexts. Nevertheless, these model-generated examples excel in their novelty and uniqueness.

In the generated example in Table~\ref{tab:exp3_example_comparison} “在山洞里放爆竹——想得深” (Setting off firecrackers in a cave—Think deeply), the figurative answer “想” (\textit{xiǎng}, to think) replaces the homophone “响” (\textit{xiǎng}, resound). This connects the deep sound of the explosion to deep thoughts.
Similarly, in “水壶烧开冒白烟——争气” (Boiling kettle emitting white smoke—Strive for success), the answer “争气” (\textit{zhēngqì}, to strive for success) is homophonous with “蒸汽” (\textit{zhēngqì}, steam). It cleverly links a physical phenomenon to a psychological state.

LLMs also make mistakes. For instance, the second example by DeepSeek-R1 in Table~\ref{tab:exp3_example_comparison}, the answer-intended “双斗” (double fight) is not a legitimate word in Chinese. This indicates that LLMs sometimes make up words when forced to be creative.

%% file: sections/related.tex
\section{Related Work}

Our investigation into the capabilities of Large Language Models (LLMs) on Chinese \xhy{} connects three critical strands of research: the tension between reasoning and memorization in model performance, the processing of linguistic humor and metaphor, and the handling of phonological ambiguity.

\subsection{Reasoning versus Memorization and Data Contamination}
A central question in current NLP research is whether LLMs solve complex tasks through genuine reasoning or by retrieving memorized patterns from their vast training corpora. \citet{biderman2023emergent} demonstrated that as models scale, they exhibit emergent memorization capabilities, often reciting long tail data verbatim. More recently, \citet{ahmed2026extracting} showed that production models have memorized substantial portions of copyrighted books, confirming the saturation of literary data in pre-training corpora.

This memorization hypothesis strongly aligns with our findings in Experiments 1 and 2, where some LLMs achieved near-perfect accuracy on existing \xhy{} but failed significantly on novel ones. Similar phenomena have been observed in mathematical reasoning. \citet{wu2025reasoningOrMemorization-fdu} revealed that the performance of models like Qwen2.5 on standard benchmarks (e.g., MATH-500) is heavily inflated by data contamination; when tested on a leakage-free dataset (RandomCalculation), performance dropped significantly, and spurious rewards failed to improve reasoning. Similarly, \citet{wang2025reasoning-or-retrieval} hypothesized that LLMs utilize two competing mechanisms—Chain-of-Thought reasoning and memory retrieval. They found that models often "hack" tasks by prioritizing retrieval over reasoning, a behavior that explains why the LLMs in our study likely retrieved the answers for dictionary-based \xhy{} rather than deriving them.

\subsection{Humor, Metaphor, and Creative Generation}
\xhy{} represents a unique intersection of humor and figurative language. Prior work on Chinese humor generation, such as the study on \textit{Xiangsheng} (Crosstalk) by \citet{li-etal-2023-crosstalk}, found that while large-scale pre-training improves generation quality, models still struggle to produce humor that matches human standards. This parallels our Experiment 3, where human evaluators consistently rated LLM-generated \xhy{} as inferior to human creations.

However, our results regarding \textit{understanding} novel \xhy{} present an interesting contrast to \citet{ichien2024llm-metaphor}, who reported that GPT-4 displays an emergent ability to interpret \textit{novel} literary metaphors, often outperforming human students. While they suggest LLMs can generalize to new figurative language, our study indicates this ability may be limited when the figurative mapping requires specific cultural schemas or rigid structural constraints, as seen in \xhy{}. The sharp 40\% accuracy drop we observed between existing and novel samples suggests that what appears to be "emergent interpretation" in other contexts might partly rely on soft-matching against memorized semantic relations, which fails when the riddle component is entirely unseen.

\subsection{Phonological Ambiguity and Puns}
Many \xhy{} rely on homophonic puns (\textit{xieyin}), adding a layer of phonological complexity. \citet{xu2024good} evaluated LLMs on English puns, identifying a "lazy generation" pattern where models fail to recognize or generate the duality of meaning required for a good pun. In the context of Chinese, \citet{ma-etal-2025-phonothink} introduced \textit{PhonoThink}, demonstrating that standard LLMs frequently struggle with Chinese phonological ambiguities without specific multi-stage training and reinforcement learning. The difficulty LLMs face in our study—particularly with novel \xhy{} that may rely on phonological cues—is consistent with these findings, suggesting that current architectures still lack robust grounding in the interplay between phonology and semantics required for deep linguistic reasoning.

%% file: sections/conclusion.tex
\section{Conclusion}
In this study, we explore the limits of LLM reasoning in a Chinese language game called \xhy{}. 
We first created X-Riddles, the first benchmark on Chinese \xhy{}, with 900 existing \xhy{} sampled from a dictionary, and more than 200 expert-written novel \xhy{} to avoid data contamination. 
We then investigated the ability to \textit{understand} and \textit{create} \xhy{} in state-of-the-art LLMs with three experiments. 
We used the difference in accuracy ($\Delta_{acc}$) between low-frequency \xhy{} and novel ones as a proxy for model memorization. 
Our results show that humans have a very low $\Delta_{acc}$ (2.9\%), suggesting genuine reasoning for both types of \xhy{}, whereas frontier Chinese models show a large mean $\Delta_{acc}$ of 23\%, indicating they may have memorized a lot of the low-frequency \xhy{} in training. 
The best frontier model (Gemini 3.1 Pro) reaches 92.6\% accuracy on the novel \xhy{} under zero-shot conditions, demonstrating genuinely strong reasoning abilities for this Chinese language game. 
However, \xhy{} created by strong LLMs (DS-R1 and GPT-o1) are rated to be less reasonable and funny compared to those created by humans. 
These findings have several implications. First, future studies need to take into consideration the data leakage issue when evaluating LLMs. Second, LLMs seem to lack phonological knowledge of Chinese characters, as they perform worse on homophonic \xhy{} than polysemous ones. Finally, humans still seem to be  better at creative language use when it comes to Chinese language games such as \xhy{}.

%% file: sections/appendix.tex
\clearpage
\appendix
\onecolumn
\section{Effects of experimental conditions on accuracy in Experiment 1}

\begin{figure*}[th!]
\centering
\includegraphics[width=1\textwidth]{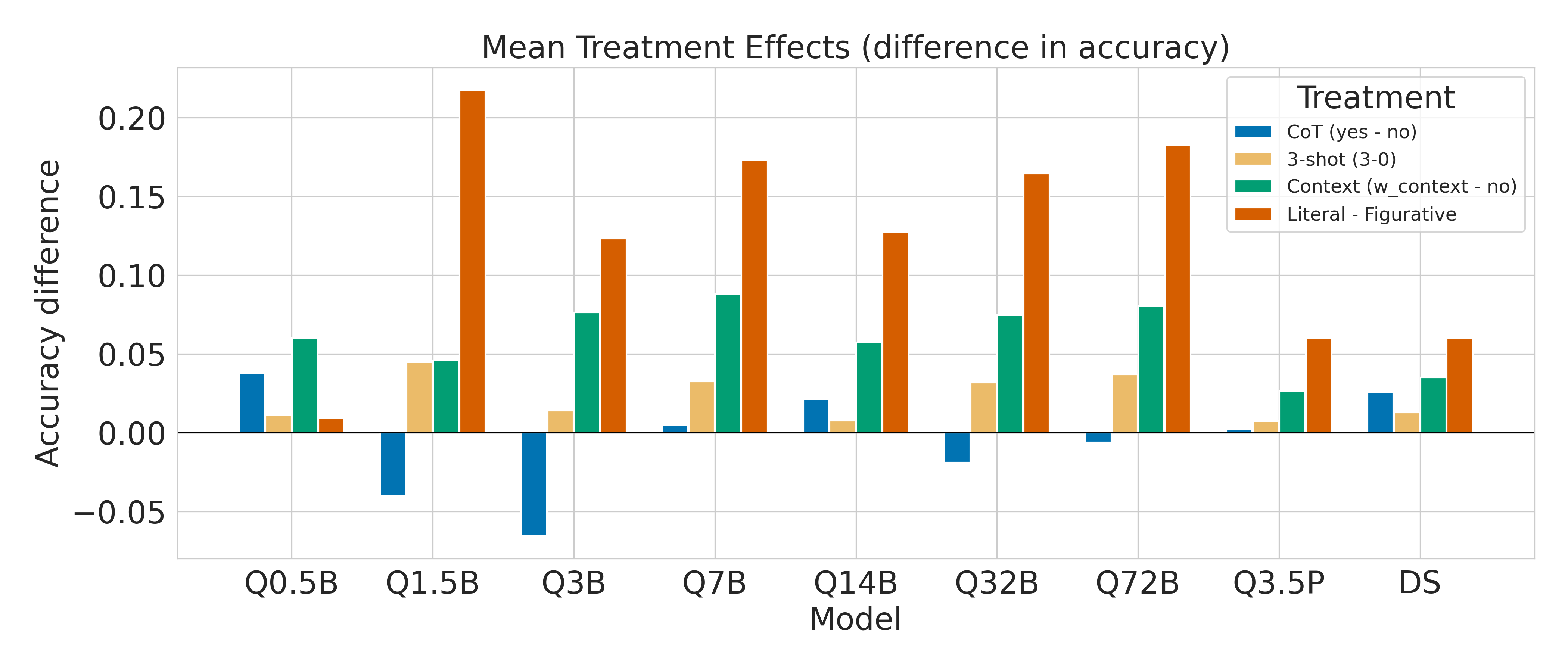}
\caption{Effects of different conditions on the overall accuracy in the multiple choice questions in Experiment 1. }
\label{fig:exp1:eval:conditions}
\end{figure*}

Effects of different conditions are presented in Figure~\ref{fig:exp1:eval:conditions}. 

\section{Examples of human-created \xhy{}}
\label{sec:app:examples:human:created:xhy}

\begin{table*}[htbp]
    \centering
    \footnotesize 
    \renewcommand{\arraystretch}{1.4} 
        \begin{tabularx}{\textwidth}{lXXXX}
        \toprule
        & \textbf{Seed} & \textbf{Human-Created} & \textbf{Human-Created} & \textbf{Human-Created} \\
        \midrule
        
        \textbf{Riddle} & 外甥打灯笼 & 外甥修城墙 & 外婆老了 & 外甥借钱 \\
        \textit{Translation} & Nephew holds lantern & Nephew builds wall & Grandma is getting old & Nephew borrows money \\
        \midrule
        
        \textbf{Ans-lit} & 照舅 & 守舅 & 依舅 & 欠舅 \\
        \textit{Pinyin} & \textit{zhào jiù} & \textit{shǒu jiù} & \textit{yī jiù} & \textit{qiàn jiù} \\
        \textit{Gloss} & light uncle & defend uncle & depend on uncle & owe uncle \\
        \textit{Translation} & light the way for uncle & protect uncle & rely on uncle & owe uncle money \\
        \midrule
        
        \textbf{Ans-fig} & 照旧 & 守旧 & 依旧 & 歉疚 \\
        \textit{Pinyin} & \textit{zhào jiù} & \textit{shǒu jiù} & \textit{yī jiù} & \textit{qiàn jiù} \\
        \textit{Gloss} & according to old & stick to old & still & guilty \\
        \textit{Meaning} & same old, as usual & conservative & as always & guilty \\    
        \bottomrule
    \end{tabularx}
    \caption{Example of three human-created \xhy{} based on a seed \xhy{}, showing the \textit{Riddle}, \textbf{Answer-literal} and \textbf{Answer-intended}, with glosses and translations. \label{tab:human:created:xhy:from:seed} }
\end{table*}

\begin{table*}[htbp]
    \centering
    \footnotesize 
    \renewcommand{\arraystretch}{1.4} 
        \begin{tabularx}{\textwidth}{lXXXX}
        \toprule
        & \textbf{Human-Created} & \textbf{Human-Created} & \textbf{Human-Created} & \textbf{Human-Created} \\
        \midrule
        
        \textbf{Riddle} & 缝纫机罢工 & 厕所里挥电蚊拍 & 吃饱了的猫 & 孙悟空穿破洞裤 \\
        \textit{Translation} & The sewing machine is down & Swinging an electric mosquito swatter in toilet & A cat with a full belly & Sun Wukong wearing ripped jeans \\
        \midrule
        
        \textbf{Ans-lit} & 不裁 & 蝇麻了 & 乐不思鼠 & 猴补 \\
        \textit{Pinyin} & \textit{bù cái} & \textit{yíng má le}‌ & \textit{lè bù sī shǔ} & \textit{hóu bǔ} \\
        \textit{Gloss} & no cutting & numb fly & too happy to care about the mouse & monkey patch \\
        \textit{Translation} & seamless & the fly is numb & too happy to miss the mouse & monkey mending clothes \\
        \midrule
        
        \textbf{Ans-fig} & 不才 & 赢麻了 & 乐不思蜀 & 候补 \\
        \textit{Pinyin} & \textit{bù cái} & \textit{yíng má le‌} & \textit{lè bù sī shǔ} & \textit{hòu bǔ} \\
        \textit{Gloss} & incompetent & endless wins  & too happy to think of Sichuan & wait list \\
        \textit{Meaning} & humble self & win big & indulge in pleasure and forget home and duty & wait list \\    
        \bottomrule
    \end{tabularx}
    \caption{Examples of human-created new \xhy{} that are \textit{not} based on existing seed \xhy{} (free style creation). \label{tab:human:created:xhy:freestyle}}
\end{table*}

        

\clearpage
\section{Prompt Templates for three Experiments}
\label{sec:app:prompt:templates}


\begin{table*}[htbp]
    \centering
    \small 
    \renewcommand{\arraystretch}{1.4} 
    \begin{tabularx}{\textwidth}{@{}X X@{}}
        \toprule
        \textbf{Original Prompt in Chinese} & \textbf{English Translation} \\
        \midrule
        【\textit{说明}】歇后语是由两部分组成的固定语句，前一部分多用比喻，像谜面，后一部分是本意，像谜底，通常只说前一部分，后一部分不言而喻。 & [\textit{Instruction}] A \textit{Xiehouyu} (allegorical saying) is a fixed phrase composed of two parts. The first part often uses a metaphor, acting like a riddle; the second part is the actual meaning, acting like the answer. Usually, only the first part is spoken, as the second part is self-evident. \\
        \addlinespace
        【\textit{任务}】你的任务是根据给出例句的上下文，以及歇后语的前半部分，选出后半部分。 & [\textit{Task}] Your task is to select the second part of the \textit{Xiehouyu} based on the context of the given sentence and the first part of the saying. \\
        \midrule
        \textbf{例子：}天上灰布悬，地下水涟涟。山区的气候，好像孙猴子的脸——\underline{\hspace{1cm}}。 & \textbf{Example:} Grey cloth hangs in the sky, water ripples on the ground. The climate in the mountains is like the Monkey King's face—\underline{\hspace{1cm}}. \\
        在上面的句子里，“\underline{\hspace{1cm}}”处应该填什么？ & What should be filled in the blank ``\underline{\hspace{1cm}}'' in the sentence above? \\
        A: 碎拾掇 \quad B: 说变就变 & A: Tidying up scraps \quad B: Changes at the drop of a hat \\
        C: 短中取长 \quad D: 两头受气 & C: Taking the long from the short \quad D: Caught in the middle \\
        \textbf{答案：} B & \textbf{Answer:} B \\
        \midrule
        \textbf{题目：}老乔的话对着哩，咱们都是一些摸牛尾巴的大老粗，啥也不懂，就敢揽那玩艺?再说今年这年成，困难比河里的石头子、砂子粒儿还多，再修水库，真成了房檐上玩把戏，\underline{\hspace{1cm}}了。 & \textbf{Question:} Old Qiao is right. We are all uneducated roughnecks who only know how to touch a cow's tail. We don't understand anything, yet we dare to take on that thing? Besides, look at the harvest this year; the difficulties are more numerous than the stones and sand in the river. If we try to build a reservoir now, it would truly be performing acrobatics on the eaves—\underline{\hspace{1cm}}. \\
        在上面的句子里，“\underline{\hspace{1cm}}”处应该填什么？ & What should be filled in the blank ``\underline{\hspace{1cm}}'' in the sentence above? \\
        A. 苦上加苦 \quad B. 豁出去 & A. Adding bitterness to bitterness \quad B. Risking it all \\
        C. 不要命 \quad D. 肚子里编成 & C. Risking one's life / Reckless \quad D. Made up inside one's belly \\
        \addlinespace
        【\textit{要求}】请直接在“答案：”后写出答案。 & [\textit{Requirement}] Please write the answer directly after ``Answer:''. \\
        \bottomrule
    \end{tabularx}
    \caption{The prompt used for Experiment 1 under the \textit{1-shot}, \textit{no CoT} condition. The model is provided with the definition, an example, and the specific question.}
    \label{tab:prompt:exp1}
\end{table*}

\begin{table*}[htbp]
    \centering
    \small
    \renewcommand{\arraystretch}{1.3}
        \begin{tabularx}{\textwidth}{@{}X X@{}}
        \toprule
        \textbf{Original Prompt in Chinese} & \textbf{English Translation} \\
        \midrule
        请解释歇后语“孔夫子搬家”的后半部分为什么是“尽是输”。 & Please explain why the answer to the riddle "Confucius move" is "always losing". \\
        \bottomrule
    \end{tabularx}
    \caption{The prompt for the \xhy{} explanation task in Experiment 2. }
    \label{tab:prompt:exp2}
\end{table*}

\begin{table*}[htbp]
    \centering
    \small
    \renewcommand{\arraystretch}{1.3}
    \begin{tabularx}{\textwidth}{@{}X X@{}}
        \toprule
        \textbf{Original Prompt in Chinese} & \textbf{English Translation} \\
        \midrule
        \textbf{定义：}歇后语是由两部分组成的固定语句，前一部分多用比喻，像谜面，后一部分是本意，像谜底，通常只说前一部分，后一部分不言而喻。 & \textbf{Definition:} A \textit{Xiehouyu} (allegorical saying) is a fixed phrase composed of two parts. The first part often uses a metaphor, acting like a riddle; the second part is the actual meaning, acting like the answer. Usually, only the first part is spoken, as the second part is self-evident. \\
        \addlinespace
        \textbf{谐音歇后语：}“谐音歇后语”是歇后语的一种，其谜底有字面义和引申义，包含一对谐音词。 & \textbf{Homophonic Xiehouyu:} "Homophonic Xiehouyu" is a type of \textit{Xiehouyu}. Its answer has both a literal meaning and a figurative (extended) meaning, containing a pair of homophones. \\
        \addlinespace
        \textbf{示例：} & \textbf{Example:} \\
        歇后语：咸菜拌豆腐——那还用言 & Xiehouyu: Pickles mixed with tofu — Needless to say (salt) \\
        谜面：咸菜拌豆腐 & Riddle: Pickles mixed with tofu \\
        字面义谜底：那还用盐 & Literal Answer: Do you still need salt? (\textit{Na hai yong yan}) \\
        引申义谜底：那还用言 & Figurative Answer: Needless to say (\textit{Na hai yong yan}) \\
        谐音字：盐/言 & Homophone: Salt (\textit{Yan}) / Say (\textit{Yan}) \\
        解释：盐： 与“言”同音相谐。采用反问语气。指心里早已明白，用不着别人多说。 & Explanation: "Salt" is homophonous with "Say". Uses a rhetorical tone. It implies the situation is already understood, and there is no need for others to say more. \\
        例句：“咸菜拌豆腐——那还用言(盐)?”小练耸起鼻孔，表示母亲的嘱咐是多余的。 & Example Sentence: "Pickles mixed with tofu — needless to say (salt)?" Xiao Lian shrugged, indicating that his mother's instructions were superfluous. \\
        \midrule
        \textbf{推理过程：} & \textbf{Reasoning Process:} \\
        第一步通过谜面的语义信息（咸菜拌豆腐），联想到用咸菜本身用盐腌过，拌豆腐时，不必再放盐，进而得到字面义谜底（那还用盐）； & Step 1: Through the semantics of the riddle (pickles mixed with tofu), one associates that pickles are already salted, so when mixing with tofu, no extra salt is needed. This leads to the literal answer (Do you still need salt?); \\
        第二步，通过字面义谜底（那还用盐），得到与之谐音的引申义谜底（那还用言），即“心里早已明白，用不着别人多说”。其中，第二步的推理用到了谐音，即例子中的“盐”和“言”。 & Step 2: Through the literal answer, one derives the homophonous figurative answer (Needless to say). Step 2 utilizes the homophone, i.e., "Salt" and "Say". \\
        \midrule
        \textbf{任务：}你的任务是根据“咸菜拌豆腐——那还用言”，写出一个与之相似的新造的“谐音歇后语”，并按照上述格式，写出“歇后语”、“谜面”、“字面义谜底”、“引申义谜底”、“解释”、“例句”。 & \textbf{Task:} Your task is to write a \textit{newly coined} "Homophonic Xiehouyu" similar to "Pickles mixed with tofu — Needless to say," based on that example. Follow the format above to write the "Xiehouyu," "Riddle," "Literal Answer," "Figurative Answer," "Explanation," and "Example Sentence." \\
        \midrule
        \textbf{新造示例（1-shot）：} & \textbf{New Creation Example (1-shot):} \\
        歇后语：咸菜汤里加酱油——察言观色 & Xiehouyu: Adding soy sauce to pickle soup — Observing words and expressions \\
        谜面：咸菜汤里加酱油 & Riddle: Adding soy sauce to pickle soup \\
        字面义谜底：察盐观色 & Literal Answer: Checking salt and observing color (\textit{Cha yan guan se}) \\
        引申义谜底：察言观色 & Figurative Answer: Observing words and expressions (\textit{Cha yan guan se}) \\
        谐音字：盐：言 & Homophone: Salt (\textit{Yan}) : Words (\textit{Yan}) \\
        解释：盐： 与“言”同音相谐。指揣度对方的话语，观察对方的脸色。 & Explanation: "Salt" is homophonous with "Words". Refers to speculating on someone's words and observing their facial expressions. \\
        例句：谈判时他不停试探对方底线，就像咸菜汤里加酱油——察言（盐）观色，生怕哪句话过了火候。 & Example Sentence: During negotiation, he kept testing the other party's bottom line, just like adding soy sauce to pickle soup — observing words (salt) and expressions, afraid that any sentence might cross the line. \\
        \addlinespace
        \textbf{注意：}确保不能输出已经存在的歇后语。确保谐音准确，即声母韵母必须都相同。 & \textbf{Note:} Ensure you do not output an existing Xiehouyu. Ensure the homophony is accurate (initials and finals must be identical). \\
        \bottomrule
    \end{tabularx}
    \caption{The prompt for the Homophonic \xhy{} generation task in Experiment 3. The model is instructed to generate a new \xhy{} by analogy, following the logic of the provided example.}
    \label{tab:prompt:exp3}
\end{table*}

\clearpage
\section{Example Errors from LLM output in Experiment 2}
\label{sec:app:example:errors:exp2}

\begin{table*}[htbp]
     \centering
     \small
    
    \begin{tabularx}{\textwidth}{@{} X p{4cm} l @{}}
        \toprule
        \textbf{Model response (excerpt)} & \textbf{Claimed Homophones (Error)} & \textbf{Model Name} \\
        \midrule
        
        
        在汉语中，“胡”与“虚”在某些方言或语境下读音相近（尤其在口语或谐音游戏中），而“须”（胡子的“须”）与“假”组合成“虚假”。 
        & \textbf{胡} (hú) \newline \textbf{虚} (xū) 
        & DS-R1 \\
        \addlinespace
        
        在中文中，“不鸣”与“无名”发音相似（“不鸣”读作 bù míng，“无名”读作 wú míng），但通过歇后语的谐音技巧，“不鸣”被转化为“无名”，意思是“没有名声”或“不出名”。 
        & \textbf{不鸣} (bù míng) \newline \textbf{无名} (wú míng) 
        & DS-R1 \\
        \addlinespace
        
        
        在中文中，“听”（tīng）和“瘾”（yǐn）发音相近，尤其是在一些方言或口语中，容易产生谐音。 
        & \textbf{听} (tīng) \newline \textbf{瘾} (yǐn) 
        & DS-R1 \\
        \addlinespace
        
        而“书宅”与“输家”谐音，听起来像“输家”， 
        & \textbf{书宅} (shū zhái) \newline \textbf{输家} (shū jiā) 
        & Kimi-k2 \\
        \addlinespace
        
        
        
        
        “补”与“候”谐音（尤其在方言或口语中可能更接近） 
        & \textbf{补} (bǔ) \newline \textbf{候} (hòu) 
        & Kimi-k2 \\
        \addlinespace
        
        如某些方言中“弦”读若“xián”，与“功”的韵母“ong”形成谐音 
        & \textbf{弦} (xián) \newline \textbf{功} (gōng) 
        & Kimi-k2 \\
        \addlinespace
        
        “杏核儿”读作 \textbf{xìng húr}... 但“杏核儿”在口语中快速连读时... 近似“\textbf{人儿苦}”（rénr kǔ）... 
        & \textbf{杏核儿} (xìng húr) \newline \textbf{人儿苦} (rénr kǔ) 
        & Qwen3-max \\
        
        \bottomrule
    \end{tabularx}

    \caption{Hallucinated Homophones in Model Explanations}
    \label{tab:homophone_errors}

\end{table*}

\begin{table*}[htbp]
    \centering
    \small
    \begin{tabularx}{\textwidth}{@{} X @{}}
    \toprule
    Dictionary Explanation:
    掸子： 一种除尘工具，多用鸡毛绑在小竹棍上做成。掸： 与“胆”同音相谐。常用于惊叹或指责人的胆子太大。 \\
    English translation for illustrative purposes: Duster (Dǎnzi): A type of tool used for dusting, typically made by tying chicken feathers to a small bamboo stick.
    The Pun: The character for "duster" (掸, dǎn) creates a homophonic pun with the character for "gallbladder/guts" (胆, dǎn).
    Meaning/Usage: This saying is commonly used to express astonishment at someone's audacity or to accuse someone of being overly bold or reckless (having "huge guts").
    \\\midrule
    \addlinespace
    Model Explanation: “旗杆上插鸡毛——好大的胆子”是一句形象生动、富有幽默讽刺意味的歇后语。它的后半部分“好大的胆子”是对前半部分“旗杆上插鸡毛”这一荒诞行为的点评，其逻辑和寓意可以从以下几个方面理解：一、字面荒谬性制造反差。旗杆：通常是高大、庄严的象征，古代用于悬挂旗帜，代表国家、军队或权威，不可轻慢。鸡毛：轻飘无足轻重之物，常被用来象征琐碎、不值一提，甚至“鸡毛蒜皮”就是形容小事。把“鸡毛”插在“旗杆”上——本应挂旗帜的庄严位置，却插了个轻飘飘、毫无分量的东西，显得极其荒唐、不合规矩。这种行为本身就需要“胆大包天”才敢做，所以自然引申出“好大的胆子”。
    …… \\ 
     English translation for illustrative purposes: The xiehouyu ``Sticking chicken feathers on a flagpole — what a huge guts!'' is a vivid folk saying characterized by humor and satire. The second part, ``what a huge guts,'' comments on the absurd action described in the first part, ``sticking chicken feathers on a flagpole.'' Its logic and implied meaning can be understood from the following aspects. First, the literal absurdity creates a strong contrast. A flagpole is typically tall and solemn; in traditional contexts, it is used to hoist flags representing the state, the military, or authority, and therefore should not be treated lightly. A chicken feather, by contrast, is light and insignificant, often symbolizing something trivial or worthless. Placing a chicken feather on a flagpole—where a flag is supposed to be raised—appears highly inappropriate and ridiculous. Such an act would require extraordinary audacity, which naturally leads to the judgment ``what enormous nerve.'' 
        ...... \\
        \bottomrule
    \end{tabularx}
    \caption{Dictionary and model explanations of the \xhy{}: 旗杆上插鸡毛——好大的掸子(胆子) \textit{Sticking chicken feathers on a flagpole — what a huge duster (guts)!}.}
    \label{tab:exp2:over-interpretation:example}
\end{table*}

\section{Measuring novelty of created \xhy{}}
\label{sec:app:novelty:measure}

Given a new \xhy{} $(r,a)$, where $r$ is the \textit{riddle} and $a$ is the \textit{answer-intended},
we concatenate them into a single string $s$.
For each existing \xhy{} $(r_i,a_i)$ in the dictionary $\mathcal{D}_{\text{old}}$,
we similarly form $s_i$. Let $\mathrm{set}(\cdot)$ map a string to the set of its \emph{distinct characters}. We compute overlap score between two \xhy{} using the Jaccard index:
\[
\mathrm{Overlap}(s,s_i)
\;=\;
\frac{|\mathrm{set}(s) \cap \mathrm{set}(s_i)|}{|\mathrm{set}(s) \cup \mathrm{set}(s_i)||}.
\]
We compute this score for all $i$ and take the maximum overlap
\[
m \;=\; \max_{i} \mathrm{Overlap}(s,s_i).
\]

If $m < 0.5$ (with $\tau = 0.5$ in our implementation), we label the new \xhy{} as \texttt{original}, and return no candidates.
Otherwise, we collect the set of \xhy{} achieving the same maximum overlap $
\mathcal{T}
\;=\;
\Bigl\{\, i \;:\; \mathrm{Overlap}(s,s_i) = m \Bigr\}$ and label the new \xhy{} as \texttt{possible\_dup}.

